# Modeling Visual Information Processing in Brain: A Computer Vision Point of View and Approach

Emanuel Diamant

VIDIA-mant, P.O. Box 933, 55100 Kiriat Ono, Israel

emanl@012.net.il

**Abstract.** We live in the Information Age, and information has become a critically important component of our life. The success of the Internet made huge amounts of it easily available and accessible to everyone. To keep the flow of this information manageable, means for its faultless circulation and effective handling have become urgently required. Considerable research efforts are dedicated today to address this necessity, but they are seriously hampered by the lack of a common agreement about "What is information?" In particular, what is "visual information" – human's primary input from the surrounding world. The problem is further aggravated by a long-lasting stance borrowed from the biological vision research that assumes human-like information processing as an enigmatic mix of perceptual and cognitive vision faculties. I am trying to find a remedy for this bizarre situation. Relying on a new definition of "information", which can be derived from Kolmogorov's complexity theory and Chaitin's notion of algorithmic information, I propose a unifying framework for visual information processing, which explicitly accounts for the perceptual and cognitive image processing peculiarities. I believe that this framework will be useful to overcome the difficulties that are impeding our attempts to develop the right model of human-like intelligent image processing.

## 1 Introduction

The explosive growth of visual information in our surroundings has raised an urgent demand for effective means for organizing and handling these immense volumes of information [1]. Because humans are known to be very efficient in such tasks, it is not surprising that computer vision designers are trying again and again to get answers for their worrying problems among the solutions that Human Visual System has developed in course of millions of years of natural evolution. Nearly half of our cerebral cortex is busy with processing visual information [2], but how it is done "in vivo" remains a puzzle for many generations of thinkers, philosophers, and contemporary scientific researchers.

Nevertheless, a working theory of human visual information processing has been established about twenty five years ago by the seminal works of David Marr [3], Anne Treisman [4], Irving Biederman [5], and a large group of their associates and followers. Since then it has become a classical theory, which dominates today in all farther developments in the field. The theory considers human visual information processing as an interplay of two inversely directed processing streams. One is an unsupervised, bottom-up directed process of initial image information pieces discovery and localization. The other is a supervised, top-down directed process, which conveys the rules and the knowledge that guide the linking and binding of these disjoint information pieces into perceptually meaningful image objects.

In modern biological vision research, this duality is referred to as perceptual and cognitive faculties of vision. In computer vision terminology, these are the low-level and high-level paths of visual information processing. Although Treisman's theory [4] definitely positions itself as "A Feature-Integration Theory", the difficulties in defining proper rules for this feature integration have impelled a growing divergence between perceptive and cognitive fields of image processing [6]. Obviously, that was a wrong and a counter-productive development, and human vision researchers were always aware of its harmful consequence [7]. For this reason, the so-called "binding problem" has been announced as a critical exploration goal, and massive research efforts have been directed to its resolution, [8]. Unfortunately, without any discernable success.

In computer vision, the situation is even more bizarre. Thus far, computer vision community was so busy with its everyday problems that there was no time to raise basic research ventures. Principal ideas as well as their possible solutions are usually borrowed from biological vision research. Therefore, following the trends in biological vision, for decades computer vision R&D has been deeply plunged into bottom-up pixel-oriented image processing. Low-level image computations have become its prime and persistent goal, while the complicated issues of high-level processing were just neglected and disregarded.

However, it is impossible to ignore them completely. It is generally acknowledged that any kind of image processing is unfeasible without incorporation into it the high-level knowledge ingredients. For this reason, the whole history of computer-based image processing is an endless saga on attempts to seize the needed knowledge in any possible way. The oldest and the most common ploy is to capitalize on the domain expert's knowledge and adapt it to each and every application case. It is not surprising, therefore, that the whole realm of image processing has been, and continues to be, fragmented according to the high-level knowledge competence of the experts in the corresponding domains. That is why we have today: medical imaging, aerospace imaging, infrared, biologic, underwater, geophysics, remote sensing, microscopy, radar, biomedical, X-ray, and so on "imagings".

The advent of the Internet, with huge volumes of visual information scattered over the web, has demolished the long-lasting custom of capitalizing on the expert's knowledge. Image information content on the Web is unpredictable and diversified. It is useless to apply specific expert knowledge to a random set of distant images. To meet the challenge, the computer vision community has undertaken an enterprise to develop the so-called Content-Based Image Retrieval (CBIR) technologies, [9], [10]. However, deprived of any reasonable sources of the desired high-level information, computer vision designers were forced to proceed in only one possible direction of trying to derive the high-level knowledge from the available low-level information pieces, [11], [12].

In doing so, computer vision designers have once again demonstrated their reliance on biological vision trends and fashions. In biological vision, a rank of theoretical and experimental work has been done in order to support and to justify this above-mentioned tendency. Two ways of thinking could be distinguished in this regard: chaotic attractors modeling [13], [14], and saliency attention map modeling [15], [16]. We will not review these approaches in details. We will only note that both of them presume low-level bottom-up processing as the most proper way for high-level information recovery. Both are computationally expensive. Both definitely violate the basic assumption about the leading role of high-level knowledge in the low-level information processing.

It will be a mistake to say that computer vision people are not aware of these discrepancies. On the contrary, they are well informed about what is going on in the field. However, they are trying to justify their attempts by promoting a concept of a "semantic gap", an imaginary gap between low- and high-level image features. They sincerely believe that they would be able to bridge it some day, [17].

It is worth to mention that all these developments – feature binding in biological vision and semantic gap bridging in computer vision – are evolving in atmosphere of total indifference to prior claims about high-level information superiority in the general course of visual information processing. Such indifference seems to stem from a very loose understanding about what is the concept of "information", what is the right way to use it properly, and what information treatment options could arise from this understanding.

## 2   Re-examining the basic assumptions

Everyone, who is not deaf, knows that we live today in the Information Age, where information is an indispensable ingredient of our life. We consume it, create it, seek it, transfer, exchange, hide, reveal, accumulate, and disseminate it – in one word: information is a remarkably important component of our life. But can someone explain me what we have in mind when the word "information" is uttered? My attempts (undertaken several years ago) to get my own answer for this question were so desperate that I was almost ready to accept the stance that information is an indefinable entity (like "space" and "time" in classical physics, e.g.). Fortunately, at the end, I have hit on an information definition fitting my visual information handling aims. It turns out that this definition can be derived from Solomonoff's theory of Inference [18], Chaitin's Algorithmic Information theory [19], and Kolmogorov's Complexity theory [20]. Recently, I have learned that Kolmogorov's Complexity and Chaitin's Algorithmic Information theory are referred as respected items of a list of seven possible contestants suitable to define what actually information is [21]. In this regard, I was very proud of myself that I was lucky to avoid the traps of Shannon's Information Theory, which is known to be useful in communication applications, but it is absolutely inappropriate for visual information explorations that I am trying to conduct. The reason for this is that Shannon's information properly describes the integrated properties of an information message, while Kolmogorov's definition is suitable for evaluation of information content of separate isolated subparts of a message (separate message objects). This is, indeed, much closer to the way in which humans perceive and grasp their visual information.

The results of my investigation have been already published on several occasions, [22], [23], [24], [25], and interested readers can easily get them from a number of freely accessible repositories (e.g., arXiv, CiteSeer (the former Research Index), Eprintweb, etc.). Therefore, I will only repeat here some important points of these early publications, which properly reflect my current understanding of the matters.

The main point is that **information is a description**, a certain alphabet-based or language-based description, which Kolmogorov's theory regards as a program that (when executed) trustworthy reproduces the original object [26]. In an image, such objects are visible data structures from which an image consists of. So, a set of reproducible descriptions of image data structures is the information contained in an image.

The Kolmogorov's theory prescribes the way in which such descriptions must be created: at first, the most simplified and generalized structure must be described. Then, as the level of generalization is gradually decreased, more and more fine-grained image details (structures) are become revealed and depicted. This is the second important point, which follows from the theory's pure mathematical considerations: image **information is a hierarchy of recursive decreasing level descriptions** of information details, which unfolds in a coarse-to-fine top-down manner. (Attention, please: any bottom-up processing is not mentioned here. There is no low-level feature gathering and no feature binding. The only proper way for image information elicitation is a top-down coarse-to-fine way of image processing.)

The third prominent point, which immediately pops-up from the two just mentioned above, is that the top-down manner of image **information elicitation does not require incorporation of any high-level knowledge** for its successful accomplishment. It is totally free from any high-level guiding rules and inspirations. What immediately follows from this, is that high-level image semantics is not an integrated part of image information content (as it is traditionally assumed). It can not be seen more as a natural property of an image. Image semantics must be seen as a property of a human observer that watches and scrutinizes an image. That is why we can say now: **semantics is assigned to an image by a human observer**. That is strongly at variance with the contemporary views on the concept of semantic information. Following the new information elicitation rules, it is impossible to continue to pretend that semantics can be **extracted from an image**, (as in [27], [28]), or should be **derived from low-level information features** via the semantic gap bridging, (as in [29], [30], and many others). That simply does not hold any more.

## 3   Computer Vision Implications

This new definition of information has forced us to reconsider our former approach to image information processing. The validity of our new assumptions and the inevitable changes in design philosophy that acceptance of these assumptions imply, have motivated us to test the issues in a framework of a visual robot design enterprise. The enterprise is aimed to creating an artificial vision system with some human-like cognitive capabilities. It is generally agreed that the first stage of such a system has to be an image segmentation stage at which the whole bulk of image pixels (image raw data) has to be decomposed into a finite set of image patches. The latter are submitted afterwards to a process of image content analysis and interpretation.

A practical algorithm based on the announced above principles has been developed and subjected to some systematic evaluations. The results were published, and can be found in [23], [24], [25]. There is no need to repeat again and again that excellent, previously unattainable segmentation results have been attained in these tests, undoubtedly corroborating the new information processing principles. Not only an unsupervised segmentation of image content has been achieved, (in a top-down coarse-to-fine processing manner, without any involvement of high-level knowledge). A hierarchy of descriptions for each and every segmented lot (segmented object) has been achieved as well. It contains the center of mass coordinates, the direction of object's main axeses, object's contour and shape depiction rules (in a system of these axeses), and other object related parameters (object related **information**), which enable subsequent object reconstruction. That is exactly what we have previously defined as information. That is the reason why we specify this information as "physical information", because that is the only information present in an image, and therefore **the only information that can be extracted from an image**. For that reason it must be dissociated from the semantic information, which (as we understand now) is a property of an external observer. Therefore it must be treated (or modeled) in accordance with specific his/her cognitive information processing rules.

What are these rules? A consensus view on this topic does not exist as yet in the biological vision theories and in the computer vision practice. So, we have to blaze our own trails. We decided, thus, to meet this challenge by suggesting a new approach based on our previously declared information elicitation principles. The preliminary results of our first attempt were published recently in [31]. As in the case of physical information, we will not repeat here all the details of this publication. We will proceed with only a brief reproduction of some critical points needed to follow up our discussion.

Human's cognitive abilities (including the aptness for image interpretation and the capacity to assign semantics to an image) are empowered by the existence of a huge knowledge base about the things in the surrounding world kept in human brain/head. This knowledge base is permanently upgraded and updated during the human's life span. So, if we intend to endow our visual robot with some cognitive capabilities we have to provide it with something equivalent to this (human) knowledge base.

It goes without saying that this knowledge base will never be as large and developed as its human prototype. But we are not sure that the requirement to be large and huge is valid in our case. After all, humans are also not equal in their cognitive capacity, and the magnitude, the content of their knowledge bases is very diversified too. (The knowledge base of aerial photographs interpreter is certainly different from the knowledge base of roentgen images interpreter, or IVUS images, or PET images). The knowledge base of our visual robot has to be small enough to be effective and manageable, but sufficiently ample to ensure robot's acceptable performance. Certainly, for our feasibility study we can be satisfied even with a relatively small, specific-task-oriented knowledge base.

The next crucial point is the knowledge (base) representation issue. To deal with it, we first of all must arrive at a common agreement about what is the meaning of the term "knowledge". (A question that usually has not a commonly accepted answer.) We state that in our case a suitable and a sufficient definition of it would be: "Knowledge is a memorized information". Consequently, we can say that knowledge (like information) must be a hierarchy of descriptive items, with the grade of description details growing in a top-down manner at the descending levels of the hierarchy.

What else must be mentioned here, is that these descriptions have to be implemented in some alphabet (as it is in the case of physical information) or in a description language (which better fits the semantic information case). Any farther argument being put aside, we will declare that the most suitable language in our case is a natural human language. After all, the real knowledge bases that we are familiar with are implemented on a natural human language basis.

The next step, then, is predetermined: if natural language is a suitable description implement, the suitable form of this implementation is a narrative, a story tale [32]. If the description hierarchy can be seen as an inverted tree, then the branches of this tree are the stories that encapsulate human's experience with the surrounding world. And the leaves of these branches are single words from which the stories are composed of. In computer vision terminology these single words are defined as nodes.

The descent into description details, however, does not stop here, and each single word can be farther decomposed into its attributes and rules that describe the relations between the attributes. At this stage the notion of physical information comes back to the game. Because the words are usually associated with physical objects in the real world, words' attributes must be seen as memorized physical information descriptions. Once derived (by a visual system) from the observable world and learned to be associated with a particular word, these physical information descriptions are soldered in into the knowledge base. Object recognition, thus, turns out to be a comparison and similarity test between currently acquired physical information and the one already retained in the memory. If the similarity test is successful, starting from this point in the hierarchy and climbing back on the knowledge base ladder we will obtain: first, the linguistic label for a recognized object, and second, the position of this label (word) in the context of the whole story. In this way, object's meaningful categorization can be acquired, a first stage of image annotation can be successfully accomplished, paving the way for farther meaningful (semantic) image interpretation.

One question has remained untouched in our discourse: How this artificial knowledge base has to be initially created and brought into the robot's disposal? The vigilant reader certainly remembers the fierce debates about learning capabilities of neural networks and other machine learning technologies. We are aware of these debates. But in our case they are irrelevant for a simple reason: the top-down fashion of the knowledge base development pre-determines that all responsibilities for knowledge base creation have to be placed on the shoulders of the robot designer.

Such an unexpected twist in design philosophy will be less surprising if we recall that human cognitive memory is also often defined as a "declarative memory". And the prime mode of human learning is the declarative learning mode, when the new knowledge is explicitly transferred to a developing human from his external surrounding: From a father to a child, from a teacher to a student, from an instructor to a trainee. There is evidence that this is not an especially human prerogative. Even ants are transferring knowledge in a similar way, [33]. So, our proposal that robot's knowledge base has to be designed and created by the robot supervisor is sufficiently correct and is fitting our general concept of information use and management.

## 4  Brain Vision Implications

Since the beginning of the computer vision age, there was a common belief that biological vision is an endless source of inspiration for the computer vision designs, and that the main ideas, which underpin contemporary computer vision implementations, are borrowed from the fruitful fields of biological vision research. In the Introduction Section of this paper, we have shown that such farfetched credits are groundless. In the course of time, only the bottom-up processing philosophy has remained as a common feature that still can be found both in biological and in computer vision designs. However, practical implementations of this philosophy have led to very different and incompatible developments: massive DSP-based parallel processing in computer vision and selective attention-based sequential processing in biological vision. This divergence is an annoying misfortune. In [25] I tried to allege that there must be a general underpinning basis able to reconcile the this day detached subdivisions. I believe that the unified information-processing framework proposed in this paper can be useful in pursuing this goal. In this regard, it is tempting to see how biological vision research can benefit from the proposed new ideas and to what extent elements of these new ideas can be discerned in the ongoing biological vision experiments.

While the mainstream of biological vision research continues to approach visual information processing in a bottom-up fashion [34], it turns out that the idea of primary top-down processing was never been extraneous to biological vision. The first publications addressing this issue are dated by the early eighties of the last century, (David Navon at 1977 [35], and Lin Chen at 1982 [36]). The prominent authors were persistent in their views on the matters, and farther research reports have been published regularly until the recent time [37], [38]. However, it looks like they all have been overlooked, both in the biological and in the computer vision. Only in the last years, a tide of new evidence has appeared that definitely advocates for the top-down processing primordiality [39], [40].

The field of cognitive vision is not ready yet to leave the traditional information processing dogmas. However, supporting evidence for a "declarative" interpretation of physical information can be already found in [41], where it is convincingly shown how a color is being "assigned" to an object.

Knowledge transfer from the outside and information description exchange in course of knowledge base building and declarative learning accomplishment have been observed not only in humans. They are ubiquitous even among far more primitive living beings: ants that learn in tandem [33] we have already mentioned in Section 3. The so-called Horizontal Gene Transfer responsible for antibiotic resistance development of bacteria [42] can also be seen as a supporting evidence for a top-down (external) knowledge base creation phenomenon.

However, the most surprising insights are still awaiting their farther clarification and confirmation. If our definition of information as a description is correct, then the current belief that a spiking neuron burst is a valid form of information representation and exchange [43] does not hold any more. Variances in spikes' heights or duty times are not an adequate alphabet to implement information descriptions of a desired complexity. We can boldly speculate that a biomolecular alphabet would be a much better and appropriate solution for such cases. Support for this kind of speculations can be derived from the recent advances in molecular biology research [44], [45]. The spikes that we observe and investigate today could be seen as a reflection of charges that are carried by the ionized parts of the molecular information messages.

This molecular description hypothesis fits very well also our new brain memory organization theory, which pretty well resembles the paradigm of computer memory use. Dendrite spines can be seen as a proper accommodation for the molecular descriptors, providing a kind of a biological hardware, a biological memory buffer, which is able to hold read/write-able information messages. Computer inspired hypotheses about "object files" [46] and "event files" [47] are repeatedly emerging in biological vision literature over the last decades. It would be interesting to notice that the concept of a biological event file fits very well also the narrative knowledge representation hypothesis proposed earlier in the Section 3.

Farther support for the idea that a complex information description can be stored in a single neuron memory cell can be seen in [48]. In this case as well, reactivation and retrieval of a memorized description resembles well the paradigm of a computer memory bank store/fetch transaction, a single_write/multiple_read memory access operation.

I am definitely excited by the options that brain vision research can gain from such a back projection of a computer vision theory (about the essence of information) on the issues of modeling visual information processing in human brain.

## 5   Some concluding remarks

In this paper, we propose a new definition of information, in particular, a definition of visual information, which is the prime point of our concern. Relying on this definition, a substantial progress has been achieved in modeling human-like intelligent image processing.

The approach outlined in this paper appears to conflict with the mainstream of research and development enterprises undertaken in the frame of European IST and USA TRECVID initiatives [49], [50]. The main points of divergence are as follows. The bottom-up way of image processing (from local details to more general forms of representation), commonly accepted today as the preferred processing alternative, is applicable only if the similarity metrics can be somehow defined a priory. Generally, that is not the case. The right way of visual information processing is a top-down (from coarse to fine) processing alternative, as it is proposed in this paper.

Relying on the bottom-up processing, the right structure of the information hierarchy can never be caught properly. That is the source of many deadfalls that the mainstream affiliates experience over and over again. For example, the unlucky attempts to solve the semantic gap problem stem from an improper placement of perceptual and cognitive vision features on the same hierarchical level. The same bottom-up preferences perpetually derail the machine-learning-based approaches for knowledge/information acquisition and handling. From a top-down processing perspective, a "machine teaching" approach, when knowledge is brought into from the outside, looks as a far more appropriate solution.

The examples mentioned above do not exhaust the list of changes that must be imposed on the mainstream projects if the new information processing rules are taken into the consideration. It is obvious that these rules are incomplete and tentative, since this is just a first step, and further research remains to be done. The enterprise that we are aimed at, is not a task for a single person or a small group of developers. It requires consolidated efforts of many interesting parties. We hope that the time for this collaboration is not far away.

## References

1. M.L. Khefri, D. Ziou, A. Bernardi, "Image Retrieval From the World Wide Web: Issues, Techniques, and Systems", *ACM Computing Surveys*, vol. 36, no. 1, pp. 35-67, March 2004.

2. D. Milner and M. Goodale, "The Visual Brain in Action", *Oxford Psychology Series,* No. 27, Oxford University Press, 1998.

3. D. Marr, "Vision: A Computational Investigation into the Human Representation and Processing of Visual Information", Freeman, San Francisco, 1982.

4. A. Treisman and G. Gelade, "A feature-integration theory of attention", *Cognitive Psychology*, vol. 12, pp. 97-136, Jan. 1980.

5. I. Biederman, "Recognition-by-Components: A Theory of Human Image Understanding", *Psychological Review*, vol. 94, no. 2, pp. 115-147, 1987.

6. L. W. Barsalou, "Perceptual symbol systems", *Behavioral and Brain Sciences*, vol. 22, pp. 577-660, 1999.

7. T. Palmeri and I. Gauthier, "Visual Object Understanding", Nature Reviews: Neuroscience, vol. 5, pp. 291-304, April 2004.

8. A. Treisman, "The binding problem". *Current Opinion in Neurobiology,* vol. 6, pp.171-178, 1996.

9. S.-F. Chang, W.-Y. Ma, A. Smeulders, "Recent Advances and Challenges of Semantic Image/Video Search", In: *Proceedings of the IEEE International Conference on Acoustics, Speech, and Signal Processing (ICASSP-2007)*. <http://staff.science.uva.nl/~smeulder/>

10. M.S. Lew, N. Sebe, C. Djeraba, R. Jain, "Content-based Multimedia Information Retrieval: State of the Art and Challenges", In *ACM Transactions on Multimedia Computing, Communications, and Applications*, February 2006.

11. A. Mojsilovic and B. Rogowitz, "Capturing image semantics with low-level descriptors", In: *Proceedings of the International Conference on Image Processing (ICIP-01)*, pp. 18-21, Thessaloniki, Greece, October 2001.

12. C. Zhang and T. Chen, "From Low Level Features to High Level Semantics", In: *Handbook of Video Databases: Design and Applications*, by Furht, Borko/ Marques, Oge, Publisher: CRC Press, October 2003.

13. K. McRae, "Semantic Memory: Some insights from Feature-based Connectionist Attractor Networks", Ed. B. H. Ross, *The Psychology of Learning and Motivation*, vol. 45, 2004. Available: http://amdrae.ssc.uwo.ca/.

14. C. Johansson and A. Lansner, "Attractor Memory with Self-organizing Input", *Workshop on Biologically Inspired Approaches to Advanced Information Technology (BioADIT 2005)*, LNCS, vol. 3853, pp. 265-280, Springer-Verlag, 2006.

15. S. Treue, "Visual attention: the where, what, how and why of saliency", *Current Opinion in Neurobiology*, vol. 13, pp. 428-432, 2003.

16. L. Itti, "Models of Bottom-Up Attention and Saliency", In: *Neurobiology of Attention*, (L. Itti, G. Rees, J. Tsotsos, Eds.), pp. 576-582, San Diego, CA: Elsevier, 2005.

17. J. Hare, P. Lewis, P. Enser, C. Sandom, "Mind the Gap: Another look at the problem of the semantic gap in image retrieval", *Proceedings of Multimedia Content Analysis, Management and Retrieval Conference,* SPIE vol. 6073, 2006. Available: http://www.ecs.soton.ac.uk/people/.

18. R. J. Solomonoff, "The Discovery of Algorithmic Probability", *Journal of Computer and System Science*, vol. 55, No. 1, pp. 73-88, 1997.

19. G. J. Chaitin, "Algorithmic Information Theory", *IBM Journal of Research and Development*, vol. 21, pp. 350-359, 1977.

20. A. Kolmogorov, "Three approaches to the quantitative definition of information", *Problems of Information and Transmission*, vol. 1, No. 1, pp. 1-7, 1965.

21. L. Floridi, "Trends in the Philosophy of Information", In: P. Adriaans, J. van Benthem (Eds.), *"Handbook of Philosophy of Information"*, Elsevier, (forthcoming). Available: http://www.philosophyofinformation.net.

22. E. Diamant, "Image information content estimation and elicitation", *WSEAS Transaction on Computers*, vol. 2, iss. 2, pp. 443-448, April 2003. http://www.worldses.org/journals/.

23. E. Diamant, "Top-Down Unsupervised Image Segmentation (it sounds like an oxymoron, but actually it isn't)", *Proceedings of the 3rd Pattern Recognition in Remote Sensing Workshop (PRRS'04)*, Kingston University, UK, August 2004.

24. E. Diamant, "Searching for image information content, its discovery, extraction, and representation", *Journal of Electronic Imaging*, vol. 14, issue 1, January-March 2005.

25. E. Diamant, "Does a plane imitate a bird? Does computer vision have to follow biological paradigms?", In: De Gregorio, M., et al, (Eds.), *Brain, Vision, and Artificial Intelligence*, First International Symposium Proceedings. LNCS, vol. 3704, Springer-Verlag, pp. 108-115, 2005. Available: http://www.vidiamant.info.

26. P. Vitanyi, "Meaningful Information", *IEEE Transactions on Information Theory*, vol. 52, No. 10, pp. 4617-4624, October 2006. Available: http://www.cwi.nl/~paulv/papers.

27. M. Naphade and T. S. Huang, "Extracting Semantics From Audiovisual Content: The Final Frontier in Multimedia Retrieval", *IEEE Transactions on Neural Networks*, vol. 13, No. 4, pp. 793-810, July 2002.

28. J. Jiten, B. Merialdo, B. Huet, "Semantic Feature Extraction with Multidimensional Hidden Markov Model", *SPIE Conference on Multimedia Content Analysis, Management and Retrieval*, January 17-19, 2006, San Jose, USA, SPIE vol. 6073, pp. 211-221, 2006.

29. Xiang Sean Zhou, T.S. Huang, "CBIR: From low-Level Features to High-Level Semantics", *Proceedings SPIE*, vol. 3974, pp. 426-431, San Jose, CA, January 24-28, 2000. Available: http://www.ifp.uiuc.edu/~xzhou2/.

30. K. Petridis, I. Kompatsiaris, M. Strintzis, S. Bloehdorn, S. Handschuh, S. Staab, N. Simou, V. Tzouvars, Y. Avrithis, "Knowledge Representation for Semantic Multimedia Content Analysis and Reasoning", *IEEE Transactions on CSVT*, vol. 15, No. 10, pp. 1210-1244, October 2005. Available: http://www.iti.gr/db.php/publications.

31. E. Diamant, "In Quest of Image Semantics: Are We Looking for It Under the Right Lamppost?", Available: http://arxiv.org/abs/cs.CV/0609003.

32. M. Tuffield, N. Shadbolt, D. Millard, "Narratives as a Form of Knowledge Transfer: Narrative Theory and Semantics", Proceedings of the 1st AKT (Advance Knowledge Technologies) Symposium, Milton Keynes, UK, June 2005.

33. N. Franks, T. Richardson, "Teaching in tandem-running ants", *Nature*, 439, p. 153, January 12, 2006.

34. T. Serre, M. Kouh, C. Cadieu, U. Knoblich, G. Kreiman, T. Poggio, "A Theory of Object Recognition: Computations and Circuits in the Feedforward Path of the Ventral Stream in Primate Visual Cortex", AI Memo 2005-036 / CBCL Memo 259, MIT, Cambridge, November 2005. Available: http://web.mit.edu/serre/publications/...

35. D. Navon, "Forest Before Trees: The Precedence of Global Features in Visual Perception", *Cognitive Psychology*, 9, pp. 353-383, 1977.

36. L. Chen, "Topological structure in visual perception", *Science*, 218, pp. 699-700, 1982.

37. D. Navon, "What does a compound letter tell the psychologist's mind?", *Acta Psychologica*, vol. 114, pp. 273-309, 2003.

38. L. Chen, "The topological approach to perceptual organization", *Visual Cognition*, vol. 12, no. 4, pp. 553-637, 2005.

39. M. Ahissar, S. Hochstein, "The reverse hierarchy theory of visual perceptual learning", *Trends in Cognitive Science*, vol. 8, no. 10, pp. 457-464, 2004.

40. C-H. Juan, G. Campana, V. Walsh, "Cortical interactions in vision and awareness: hierarchies in reverse", *Progress in Brain Research*, vol. 144, pp. 117-130, 2004.

41. T. Hansen, M. Olkkonen, S. Walter & K. Gegenfurtner, "Memory modulates color appearance", *Nature Neuroscience*, vol. 9, no. 11, pp. 1367-1368, November 2006.

42. J. Lawrence, H. Hendrickson, "Lateral gene transfer: when will adolescence end?", *Molecular Microbiology*, vol. 50, no. 3, pp. 739-749, 2003.

43. K. Diba, C. Koch, I. Segev, "Spike propagation in dendrites with stochastic ion channels", *Journal of Computational Neuroscience*, vol. 20, pp. 77-84, 2006.

44. E. Kandel, "The Molecular Biology of Memory Storage: A Dialogue Between Genes and Synapses", *Science*, vol. 294, pp. 1030-1038, 2 November 2001.

45. A. Routtenberg, J. Rekart, "Post-translational protein modification as the substrate for long-lasting memory", *TRENDS in Neurosciences*, vol. 28, no. 1, pp. 12-19, January 2005.

46. D. Kahneman, A. Treisman, B. Gibbs, "The reviewing of object files: Object-specific integration of information", *Cognitive Psychology*, vol. 24, pp. 175-219, 1992.

47. B. Hommel, "Event files: Feature binding in and across perception and action", *TRENDS in Cognitive Sciences*, vol. 8, no. 11, pp. 494-500, 2004.

48. S. Waydo, A. Kraskov, R. Quiroga, I. Freid, and C. Koch, "Sparse Representation in the Human Medial Temporal Lobe", *The Journal of Neuroscience*, vol. 26, no. 40, pp. 10232 - 10234, October 4, 2006.

49. European IST Research (2005-2006): Building on Assets, Seizing Opportunities. Available: http://europa.eu.int/information_society/.

50. P. Over, T. Ianeva, W. Kraaij, A. Smeaton, "TRECVID 2005 – An Overview", Available: http://www.uv.es/~tzveta/.